\documentclass{article}
\usepackage{arxiv}

\usepackage[utf8]{inputenc} 
\usepackage[T1]{fontenc}    
\usepackage{hyperref}       
\usepackage{url}            
\usepackage{booktabs}       
\usepackage{amsfonts}       
\usepackage{nicefrac}       
\usepackage{microtype}      
\usepackage{lipsum}
\usepackage{amsmath}
\usepackage{graphicx}

\usepackage{subfig}

\title{\textbf{Seamless Copy Move Manipulation in Digital Images}}

\author{
  Tanzila Qazi \\
  Hazara University,\\
  Mansehra 21120, Pakistan\\
   \AND
  Mushtaq Ali \\
  Hazara University\\
  Mansehra 21120, Pakistan.\\
  \And
  Khizar Hayat \\
  University of Nizwa,\\
  Sultanate of Oman \\
   \texttt{khizar.hayat@unizwa.edu.om}\\
}

\begin{document}
\maketitle
\begin{abstract}\normalsize
The importance and relevance of digital image forensics has attracted researchers to establish different techniques for creating as well as detecting forgeries. The core category in passive image forgery is copy-move image forgery that affects the originality of image by applying a different transformation. In this paper frequency domain image manipulation method is being presented.The method exploits the localized nature of discrete wavelet transform (DWT) to get hold of the region of the host image to be manipulated. Both the patch and host image are subjected to DWT at the same level $l$ to get $3l + 1$ sub-bands and each sub-band of the patch is pasted to the identified region in the corresponding sub-band of the host image. The resultant manipulated host sub-bands are then subjected to inverse DWT to get the final manipulated host image. The proposed method shows good resistance against detection by two frequency domain forgery detection methods from the literature. The purpose of this research work is to create the forgery and highlight the need to produce forgery detection methods that are robust against the malicious copy-move forgery. 
\end{abstract}



\section{Introduction}
\label{sec:Intro 1}
Refer to it by any name - image manipulation, composition, editing, tampering, forgery or fakery- the ultimate victim is the integrity and authenticity of the image. And the usage spectrum is broad, with aesthetics on one extreme and malicious intents (like blackmailing and character assassination) on the other. The readily available software like Adobe Photoshop, GIMP, or even XnView - has further escalated the matter. No matter, how noble your intentions are while introducing any innovation to manipulate images, the stakes of negativity are always high. And the burden to deal with all such negativity shifts to the forensic analyst. Aptly described as "arm’s race" in~\cite{1}, this competition between the manipulator and forensic analyst may never end.

As “a picture is worth a thousand meanings”, authenticity and trustworthiness of images are both legally and socially very important. For security purposes, several approaches have been developed, broadly categorized as active and passive methods. Research in passive methods has been getting increasing attention in the community because of the limitations in the active counterparts, especially its reliance on watermarks: It should be embedded at the time of image acquisition that requires a specially equipped camera or devices and, secondly, most of the watermarks degrade the image quality while manipulating the image for the insertion of watermark or related processing~\cite{2,3}.

Blind image forensics deal with only the image under investigation and that too without expecting any additional side information. It has thus a vital role in many areas including medical imaging, news reporting, criminal or court room investigation, insurance claim investigation, sports, intelligence services and many others~\cite{1,2,3}. In passive approaches, the image is forged through some sophisticated software or image manipulation techniques, where the goal is imperceptibility as well as not to leave any observable traces indicative of fakery, like motion blur, broken edges or edge inconsistencies.

In this paper, we propose a passive copy-move image manipulation method that exploits the localized nature of the Discrete Wavelet transform (DWT). The idea is to pass the image to the frequency domain and subject each of the resultant individual sub-bands to copy/move image manipulation. In doing so, we transform both the patch and the host image to DWT domain and try to paste the sub-bands of the patch on corresponding sub-bands of the host. This is followed by the inverse DWT to get the manipulated image. By doing so, we are trying to achieve the homogeneity between the patch and its new environment in the host; the inverse DWT operation should be vigorous enough to ensure the required scrambling in order to suppress any side effects therein. The result should be the seamless manipulation of the host image. To ascertain that, we have chosen two frequency domain forgery detection methods from the literature \cite{4,5} as reference. The results suggest that these two methods were not that successful to detect the manipulation.
The rest of the paper is arranged as follows. Section 2 presents a concise survey of the related literature. This is followed by the presentation of the proposed method in Section 3. The simulation results are illustrated in Section 4 in somewhat detail. Section 5 concludes the paper.

\section{ LITERATURE REVIEW}
  In this section the image forgery and forgery detection methods are presented, the purpose of this research work is to generate an image forgery that has no visual clues or tampering evidence and further on we need to the check the strength of the forged image created by our own proposed method with the existing state of the art methods.

Image manipulation or image forgery encompasses any technique that may be used to manipulate an image \cite{1}. It may be carried out using either active or a passive approaches \cite{2,3}. The active approaches are mostly concerned with the data hiding techniques, such as digital watermarks/signatures, wherein prior information is considered essential and integral to the process. The passive or blind approaches do not require any prior information about the original image [3,6,7] and the analyzer has just the final product at his/her disposal. A forensic expert will prefer the term image forgery over image manipulation and would classify it as \cite{8, 9,21}:
\begin{enumerate}
\item Copy/move forgery or cloning where the patch comes from the host image, detection is relatively hard because source and destination image is same, color and noise is same for the rest of image
\item Image splicing where the patch comes from a different image than the host, detection is comparatively easy because source and destination is from different images or set of images,
\item Image retouching which encompasses a wide spectrum of techniques to enhance the visual appearance of the image in hand, and is the least pernicious type of forgery, widely used in magazine photography editors.
\end{enumerate}
In technical terms the former two, i.e. copy/move and splicing, form the basis of what is called object transferring. Many object transfer techniques are there~\cite{1} but the following are the most popular:
\begin{itemize}
\item Cut-Out wherein the patch boundaries should be well-defined and the objective is to make the  con- tours seamless in a variety of ways, e.g. ”RepSnapping”~\cite{10}, ”Intelligent Scissors”~\cite{11} etc. Such manipulations don’t care about the original environment of the patch and are therefore easy to detect.
\item Alpha Matting refers to soft extraction of a foreground~\cite{12} and is somehow similar to cut-out but makes use of alpha-transparency adjustments between the origin and destination images  to  dilute the boundaries.
\item Gradient Domain techniques with the goal being to blend the gradient of the patch with that of the host. Among these perhaps the most popular ones are those based on the interpolations through Poisson equations for gradient match; a technique referred   to as Poisson Editing~\cite{13} or Seamless Cloning~\cite{14}.
\item Laplacian pyramids have also been employed in many works during the blending process \cite{15, 16}. \end{itemize}

As of DWT, Hayat \textit{et al.}~\cite{17} present two transform domain methods to seamlessly stitch satellite image tiles of heterogeneous resolutions. One is local and each constituent DWT domain tile of the view is treated at sub-band level with horizontal, vertical, and radial smoothing, on the basis of its locale in the tessellation. The second method assumes the view field to be of a sliding window containing parts of the tiles from the heterogeneous tessellation. The window is subjected to DWT domain mosaic king as well as smoothing. The last step in both the methods is overall inverse DWT.

Except the image manipulation by the easily available software, in~\cite{27} a forgery method is proposed for an experiment on the detection techniques that shows good resistance to forgery detection. In this method, the patch image mask is produced from the host image and further on pasted on the host image to get forged image.

In~\cite{7}, the process of image forgery is described as: selection of the region of interest (ROI), transformation and composition of image fragments and some necessary post-processing on the final image. The process usually begins with extracting the fragments and then fusing the transformed image fragments into another image using different techniques, such as matting/pasting for coherent looking composition. Visually the method claims to produce no tampering evidence in the face of the existing techniques in forgery detection. In~\cite{26} different technologies and tampering techniques are described for digital images that are difficult to detect. According to the authors, the problem of establishing the authenticity and reality of digital photography have become more complex and challenging.

The literature is replete with the surveys on forgery detection literature~\cite{2,3,24}. It will not be out of place to discuss some of the frequency domain methods, with special reference to the discrete cosine transform (DCT) and DWT. The method in~\cite{18} divides the image into overlapping blocks and computes the DCT coefficients. By using the signs of the DCT coefficients, binary feature vectors are created. The latter are matched using the coefficient of correlation. The method in \cite{19} employs scale invariant feature transform (SIFT) in combination to DCT.

DWT is a popular transform for its localized nature and the ability to compact most of the image information into the lowest energy sub-band that is dyadically reduced in size proportional to the image. Hence, rather than the suspect image, its lowest energy sub-band can be subjected to forensic analysis to reduce the complexity – a level-2 sub-band would have sixteen times less coefficients to analyze. In addition, DWT may enable the extraction of very good and robust features for comparisons. A DWT based method~\cite{20}, first, exhaustively searches for the identification of matching blocks and then uses phase correlation for the detection of the copied region. However, the technique gives poor results if the copied region is slightly scaled or rotated. In~\cite{21}, pixel matching and DWT techniques are utilized to reduce the dimensions. Moreover, phase correlation is used for the detection steps in the copied and pasted regions. To improve the forgery localization, mathematical morphology is employed for the connected regions. The above mentioned technique has low complexity and exhibits robustness against the post processing of the copied regions. However, the performance depends on the scene of the copy/move image. 

Another copy-move forgery detection algorithm for color images is on the basis of sensor pattern noise (SPN)~\cite{22}. Pattern noise is extracted by using the wavelet based Wieners denoising filter. The features are selected on the basis of signal to noise ratio, information entropy, variance of pattern noise and average of energy gradient of the extracted image. This method is shown to be robust against the geometric transformation (rotation and scaling), noise and JPEG compression. The technique in~\cite{23} is based on DWT and DCT Quantization Coefficients Decomposition (DCT-QCD). The method exhibits accuracy but does not show robustness against rotation and scaling

In \cite{25}, to increase the efficiency of detection algorithm, a hybrid approach is proposed in which block based and keypoint based feature are extracted. Fourier Mellin transform is used for extraction of feature in block based technique whereas the SIFT is used for the keypoint feature extraction. Image is first divided into the texture and smooth regions, keyppoint features are extracted from the textured regions and FMT is applied on the smooth region.extracted features are used to find the duplicated blocks or regions. This technique is comparatively less time demanding and work well in the presence of geometrical transformation. 

\section{THE PROPOSED METHOD}
\begin{figure}[ht]
    \centering
    \includegraphics[width=6cm, keepaspectratio=false]{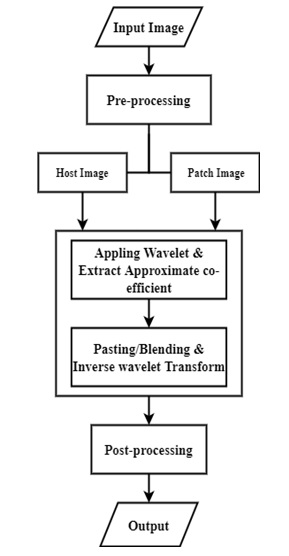}
    \caption{Block Diagram Representation of Frequency Domain Manipulation}
    \label{fig1t}
\end{figure}
A block diagram outlining the proposed method is shown in Fig.~\ref{fig1t}. The method involves the following steps:
\subsection{Preprocessing}
Before treating the image in the frequency domain some preprocessing is inevitable. These 
include the cropping of the patch that is extracted from the original image and suppressing any noise in both the host image (call it image A) and the patch (call it image B) via smoothing. Before passing to DWT domain, better convert  both  the  host  and  the patch from of RGB to YCbCr to color transform domain, in order to facilitate the manipulation of image in subsequent steps. 

\subsubsection{Applying wavelet transformation}
Apply a level $l$ DWT to each of  the YCbCr  components  of  both  A and B, separately, to get $3l+1$  sub-bands  for  each of the YCbCr components
. The size of a given sub-band is a dyadic fraction of the image size. For example if $l = 1$, then for a square image of dyadic size n × n, we get four sub-bands (LL, LH, HL and HH), each of size n/2 × n/2. The LL sub-band is the lowest frequency sub-band containing most of  the image’s energy.
    
 \subsubsection{Pasting or Blending and applying the inverse DWT}
While keeping into account the correspondence of both the components and their sub-bands, paste each sub-band of B to the identified place in the corresponding sub-band of A. This pasting may be carried out just by simple cut-out or alpha matting or even using gradient transfer, like Poisson image editing, as elaborated in Section 2. Apply level $l$ inverse DWT to the blended sub-bands from the last step to get the Y,Cb and Cr components of wavelet transformed manipulated image. 

\subsubsection{Post-Processing}
 In the last step combine all the subbands of Y,Cb and Cr to get the transformed YCbCr image. The resultant tampered image (say A’) is obtained by passing from the YCbCr back to the RGB domain.
\begin{figure*}[!htbp]
\centering
\subfloat[Transforming to DWT domain]{\includegraphics[width=7.5cm]{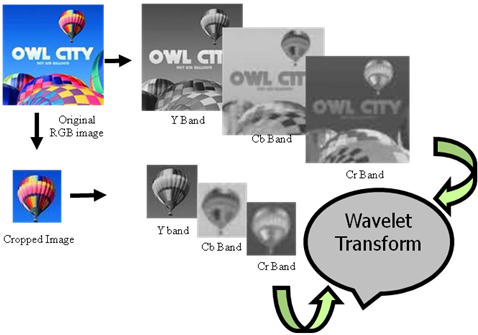}}\qquad
\subfloat[Copy/Paste and subsequent IDWT]{\includegraphics[width=7.5cm]{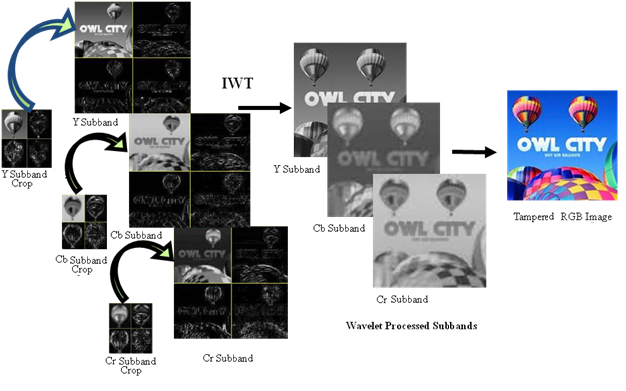}}\\
\caption{Pictorial Representation of DWT Domain Tampering.}
\label{fig14}
\end{figure*}

The overall process for this specific manipulation is graphically illustrated in Fig.~\ref{fig14}.
First, we crop out a small region (the balloon) from the original image to serve as a patch to be pasted at a predetermined position in the host image. Note that since we are doing copy/move manipulation, the original image is also the host image. As one can see in Fig.~\ref{fig14}.a, both the patch and the host are converted to YCbCr domain and subsequently each of the resultant component is subjected to DWT. In this particular case, as can be observed from Fig.~\ref{fig14}.b, only a single level ($l=1$) DWT is employed, and hence are four sub-bands each for every YCbCr component of both the host and the patch. Each sub-band of the patch is pasted to its corresponding host sub-band of a given component, to get the forged DWT domain Y, Cb and Cr components, as shown in Fig.~\ref{fig14}.b. Application of inverse DWT results in the Y, Cb and Cr components of the forged image. These three components combine to give the final forged RGB image shown in Fig.~\ref{fig2}.c.  

The whole idea behind using DWT domain pasting and subsequent inverse DWT is to dilute any artifacts that may result, especially, along the contours of the patch in its new environment in the host. We believe that inverse DWT has the capacity to smooth such artifacts.

\section{EXPERIMENTAL RESULTS}
We have applied our method to a set of images from various sources from internet and the results were interesting when observed by inspection and against two state of the art methods from the literature. This section presents first a various sources of data set, next section elaborate evaluation metrics and finally discuss the working of the proposed method and the overall results on the existing state of art techniques.  

\subsection{Data set}
Our proposed technique forged images data set consist of 30 images that are visually imperceptible and harder to detect the forged regions. A set of images have been taken from the following url given in the table. 
\subsection{EVALUATION METRICS}
The effectiveness of the forgery detection methods are usually gauged by the two measures, viz. the detection accuracy ($r$) and the false detection rate or FDR ($w$). These are computed by the following equations:
\begin{equation} 
\label{eq:05}
r=\frac{\mid R \cap D \mid}{\mid R \mid}\\  
\end{equation} 
\begin{equation} 
\label{eq:06}
w=\frac{\mid F - D \mid}{\mid R \mid}\\  
\end{equation} 
 Where $R$ represents the actual tampered area, $D$ is the detected area and $F$ is the falsely detected area.
 
\subsection{Result Evaluation}
\begin{figure*}[!htbp]
    \centering
    \subfloat[Original/Host Image]{\includegraphics[width=4.0cm]{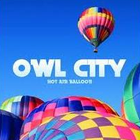}}\qquad
    \subfloat[Patch Image]{\includegraphics[width=2.5cm]{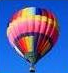}}\qquad
    \subfloat[Forged Image]{\includegraphics[width=4.0cm]{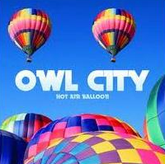}}\\
    \caption{A simulation example of copy/move forgery.}
    \label{fig2}
\end{figure*}

\begin{figure*}[!htbp]
    \centering
    \subfloat[Y]{\includegraphics[width=4.0cm]{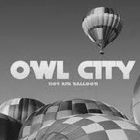}}\quad
    \subfloat[Cb]{\includegraphics[width=4.0cm]{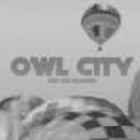}}\quad
    \subfloat[Cr]{\includegraphics[width=4.0cm]{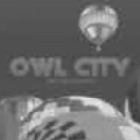}}\\
    \caption{Luminance/Chrominance components of the original image}
    \label{fig3}
%
    \includegraphics[width=6cm,height=2.0cm, keepaspectratio=false]{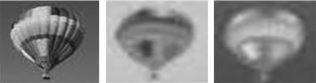}
    \caption{Luminance/Chrominance (YcbCr) components of the patch}
    \label{fig4}
\end{figure*}

For demonstration purposes we take one example from the dataset. Fig.~\ref{fig2} shows the example original image (A) and the corresponding patch image (B) taken from the host image.
The images are subjected to Component Transform in the shape of YCbCr to get the Y, Cb and Cr bands for both the host (Fig.~\ref{fig3}) and the patch (Fig.~\ref{fig4}).

\begin{figure*}[!htbp]
    \centering
     \subfloat[Original Image]{\includegraphics[width=13cm]{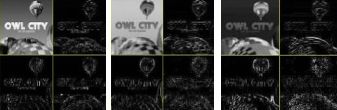}}\\
    \subfloat[Patch image]{\includegraphics[width=7.4cm]{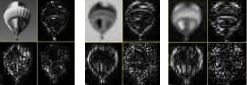}}\\
    \caption{Level-1 Wavelet transformed YCbCr components}
    \label{fig5}
    \end{figure*}

Application of level-1 DWT results in 4 subbands for each A and B, as shown in the Fig~\ref{fig5}.
    
\begin{figure*}[!htbp]
    \centering
    \subfloat[Y]{\includegraphics[width=4.0cm]{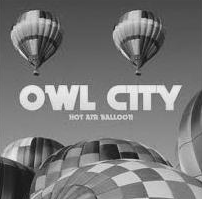}}\quad
    \subfloat[Cb]{\includegraphics[width=4.0cm]{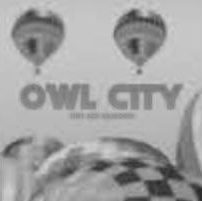}}\quad
    \subfloat[Cr]{\includegraphics[width=4.0cm]{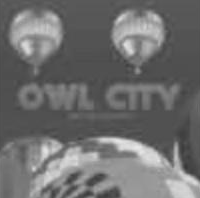}}\\
    \caption{Luminance/Chrominance components of the forged image}
    \label{fig6}
\end{figure*}

After pasting the patch subbands in the corresponding sub-bands of host, inverse DWT is applied to get the YCbCr images shown in the Fig~\ref{fig6}.

Applying inverse Component Transform to YCbCr results in the RGB image (say A') shown in Fig \ref{fig2}.c.

\begin{figure*}[!htbp]
    \centering
    \subfloat[Original]{\includegraphics[width=4cm]{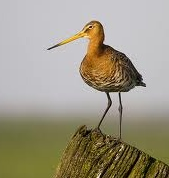}}\quad
    \subfloat[Forged]{\includegraphics[width=4cm]{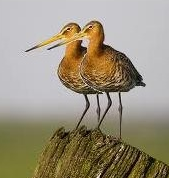}}\quad
    \subfloat[Original]{\includegraphics[width=4cm]{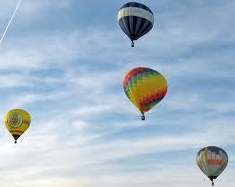}}\quad
    \subfloat[Forged]{\includegraphics[width=4cm]{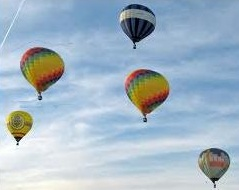}}\\
    \caption{Example Results.}
    \label{fig8}
\end{figure*}

As can be seen from the simulation result above, it is hard to discern the tampering with a naked eye, especially in the absence of the original. For illustration purposes we are showing results of two more examples in Fig.~\ref{fig8}. But one cannot solely rely on subjective results and it is therefore imperative to test the effectiveness of the proposed methods against some methods from the literature. With this in perspective, we have chosen two methods, for the sake of comparison, in order to judge the effectiveness of the proposed method. The first method, by Mahmood et al.~\cite{4}, will be hereafter called Mahmood’s method. The second method is by Meena and Tyagi~\cite{5} and will be hereinafter called Meena’s method.

\subsection{Mahmood's method}
This method detects image forgery by feature matching using Tchebichef moments of the investigated image. Image is segmented into the overlapping blocks and Tchebichef moments are computed for every block. For further dimension reduction each block is segmented into non-overlapping block and SVD  (Singular value Decomposition)  is applied to get the reduced size  feature vectors which are then sorted lexicographically in order to bring the similar vectors closest to each other. In the end morphological processing is applied to produce the result and is mapped on the basis of a preset threshold value.
This technique is claimed to be capable of unveiling the single and multiple copy move forgeries in the presence of post processing i.e brightness change, color reduction, contrast adjustment, compression and blurring. 

\subsubsection{Meena’s Method}
 In Meena’s method, the suspect image is divided into fixed sized overlapping blocks and each block is subjected to Tetrolet transform and low pass coefficients and high pass coefficients are extracted from each block. This is followed by lexicographically sorting the four features of each block in order to check the similarity measure on the basis of a threshold value. The method is claimed to be robust against the small and multiple forgeries and is applicable even if the image is scaled or passing through some post processing like blurring rotation and adjustment or brightness. Applicability of any copy/move forgery detection algorithm depends on the requirement that it should detect the duplicated regions which are either natural or forged.

\begin{figure*}[ht]
    \centering
    \subfloat[Forged]{\includegraphics[width=4cm, height=4cm]{media/balloon_fg.png}}\quad
    \subfloat[Mahmood]{\includegraphics[width=4cm, height=4cm]{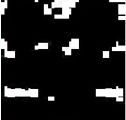}}\quad
    \subfloat[Meena]{\includegraphics[width=4cm, height=4cm]{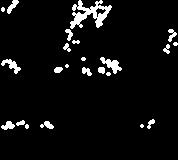}}\\
    \subfloat[Forged]{\includegraphics[width=4cm, height=4cm]{media/bird2_fg.png}}\quad
    \subfloat[Mahmood]{\includegraphics[width=4cm, height=4cm]{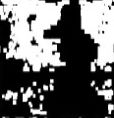}}\quad
    \subfloat[Meena]{\includegraphics[width=4cm, height=4cm]{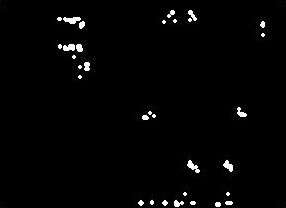}}\\
    \caption{Forgery detection results by the two benchmark methods.}
    \label{fig9}
\end{figure*}

The Fig \ref{fig9} demonstrates the detection accuracy of the two benchmark methods with two representative examples; for comparison purposes we have also included the forged images as Fig \ref{fig9}.a and Fig \ref{fig9}.d. The experimental results in Fig \ref{fig9}.b and Fig \ref{fig9}.e expose that Mahmood’s method only detected the naturally similar blocks from the forged image and gave no clues about the forged regions. Meena's method also shows a similar detection accuracy as can be seen in Fig \ref{fig9}.c and Fig \ref{fig9}.f. The method relies on plain copy move forgery i.e a part of image is copied or replicated in a same image without applying any type of post processing operation. As the image shown in Fig \ref{fig9}.c and Fig \ref{fig9}.f, the detection accuracy stands at almost zero and gives more potency to forgery detection algorithms. Therefore, the results on Meena’s methods reveal that accuracy rate of the forged regions is almost untraceable; only natural similarity is detected.

\subsubsection{Discussion}
 The average results over the set of all test images manipulated by the proposed method after subjecting to the two reference detection methods are tabulated in the form of Table 2. With Mahmood’s method the average detection accuracy is as low as 9.62 \% with a Standard deviation (sigma) of 19.81. Meena method is even worse and the accuracy is distributed  with a mean of 12.37 \% and standard deviation of 22.17.  The FDR is comparatively high for both the methods and cannot be rejected as insignificant. Thus Mahmood’s method fails to detect the forged regions and may only detect the natural similar blocks in the tampered image. Similarly the detection results of the Tatrolet transform (Meena method) are also not enviable.
%


\section{CONCLUSION AND FUTURE WORK}

By tampering images in the DWT domain and subsequently applying the inverse DWT, we were able to get robust results. The presumption that the inverse DWT has enough potential to do away with artifacts/side effects resulting from any manipulation proved to be valid at least with copy/move forgeries. As the experiments suggest, the two detection methods were not that successful in zeroing on over tampered areas for that reason, there is need to improve and refine the forgery detection methods.  As a future perspective, investigations can be carried out to combine the proposed method with state of the art image compositing techniques, especially the Gradient based methods - like Poisson image editing.

\end{document}